\documentclass{article}

\usepackage[final]{nips_2017}

\usepackage[utf8]{inputenc} % allow utf-8 input
\usepackage[T1]{fontenc}    % use 8-bit T1 fonts
\usepackage{hyperref}       % hyperlinks
\usepackage{url}            % simple URL typesetting
\usepackage{booktabs}       % professional-quality tables
\usepackage{amsfonts}       % blackboard math symbols
\usepackage{nicefrac}       % compact symbols for 1/2, etc.
\usepackage{microtype}      % microtypography
\usepackage{graphicx}  %Required
\usepackage{subcaption}
\usepackage{amsmath}

\begin{document}
% The file aaai.sty is the style file for AAAI Press 
% proceedings, working notes, and technical reports.
%
%\title{Semi-supervised Deep Clustering on Objects}
%\title{Weakly-supervised deep clustering for semantic object retrieval}
%\title{Weakly-supervised object category learning and retrieval}
\title{Object category learning and retrieval with \\weak supervision}
\author{Steven Hickson, Anelia Angelova, Irfan Essa, Rahul Sukthankar\\
Google Brain / Google Research\\
(shickson, anelia, irfanessa, sukthankar)@google.com}
%\author{Anonymous Authors}
\maketitle
\begin{abstract}
We consider the problem of retrieving objects from image data and learning to classify them into meaningful semantic categories with minimal supervision. To that end, we propose a fully differentiable unsupervised deep clustering approach to learn semantic classes in an end-to-end fashion without individual class labeling using only unlabeled object proposals. The key contributions of our work are 1) a kmeans clustering objective where the clusters are learned as parameters of the network and are represented as memory units, and  2) simultaneously building a feature representation, or embedding, while learning to cluster it.
This approach shows promising results on two popular computer vision datasets: on CIFAR10 for clustering objects, and on the more complex and challenging Cityscapes dataset for semantically discovering classes which visually correspond to cars, people, and bicycles. 
Currently, the only supervision provided is segmentation objectness masks, but this method can be extended to use an unsupervised objectness-based object generation mechanism which will make the approach completely unsupervised.
\end{abstract}

\section{Introduction}

Unsupervised discovery of common patterns is a long standing task for artificial intelligence as shown in \cite{Barlow1989unsupervised,bengio2012unsupervised}.
Recent deep learning approaches have offered major breakthroughs in classification into multiple categories with millions of labeled examples (e.g.~\cite{krizhevsky2009learning,szegedy2015going,he2016deep} and many others). These methods rely on a lot of annotated data for training in order to perform well. Unfortunately, labeling is an inefficient and expensive progress, so learning from unlabeled data is desirable for many complex tasks. At the same time, much of human knowledge and learning is obtained by unsupervised observations~\cite{Grossberg19943d}. 

The goal of this work is to show that semantically meaningful classes can be learned with minimal supervision. Given a set of objectness proposals, we use the activations of foreground objects in order to learn deep features to cluster the available data while simultaneously learning the embedding in an end-to-end manner.
More specifically, we propose a differentiable clustering approach that learns better separability of classes and embedding. 
The main idea is to store the potential cluster means as \emph{weights} in a neural network at the higher levels of feature representation. This allows them to be \emph{learned} jointly with the potential feature representation.
This differentiable clustering approach is integrated with Deep Neural Networks (e.g.~\cite{szegedy2015going}) to learn semantic classes in an end-to-end fashion without manual class labeling. 

The idea of doing this `end-to-end' is that gradient descent can not only learn good weights for clustering, it can also change the embedding to allow for better clustering without the use of labels. We see that this leads to better feature representation.
Our results show also that different object categories emerge and can later be retrieved from test images never before seen by the network, resulting in clusters of meaningful categories, such as cars, persons, bicycles.

In this work we use given segmentation objectness masks, which are candidate objects without labels. This can be extended by using an independent objectness-based object generation mechanism~\cite{pathak2017learning, faktor2014} or by using unsupervised motion segmentation in videos or structure from motion~\cite{sfmnet17}.

\section{Related Work}

Unsupervised learning (\cite{Barlow1989unsupervised}) and unsupervised deep learning (\cite{bengio2012unsupervised},~\cite{bengio2012deep},~\cite{bengio2009learning}) are central topics to Machine Learning.  
Unsupervised deep learning has been shown to improve results on classification tasks per \cite{erhan2010does}, especially given small datasets and complicated high dimensional data such as video. This has been explored by many representations including sequence to sequence learning and textual representations (\cite{radford2017learning}, \cite{ramachandran2016unsupervised}). 

Our work focuses on unsupervised deep learning for discovering visual object categories. This has also been shown to improve results such as in \cite{doersch2017multi}. Unsupervised discovery of visual objects has been a large topic of interest in computer vision (\cite{sivic2005discovering,russell2006using,singh2012unsupervised,bach2003spectral,kwak2015unsupervised,pathak2017learning}).

Building specialized, deep embeddings to help computer vision tasks is also a popular approach~ such as in \cite{agrawal2015learning}. Transfer learning from supervised tasks has proven to be very successful. Further, ~\cite{agrawal2015learning} propose learning the lower dimensional embedding through unsupervised learning and show improved performance when transfered to other supervised tasks.

Despite the popularity of building different embeddings, there is little work investigating the use of clustering to modify the embedding in an end-to-end deep learning framework. \cite{bottou1995convergence} investigate a differentiable version of the kmeans algorithm and examine its convergence properties. 
Our work focuses on {\it learnable} feature representations (instead of fixed ones as in~\cite{bottou1995convergence}) and introduces memory units for the task.

\section{Unsupervised deep clustering}
\label{sec:approach}
%\subsection{}

Our unsupervised deep clustering is inspired by \cite{bottou1995convergence}, who consider  differentiable clustering algorithms. 

We differ from this approach because the features we cluster also change with backpropogation.
In our work, we add a kmeans-like loss that is integrated end-to-end. Our idea is to store the potential cluster means as \emph{weights} in the network and thus have them be \emph{learned}.

The proposed clustering is done simultaneously while building an embedding. Given information of a potential object vs background (binary labels), clustering in a differentiable way provides a better embedding for the input data. We show that this method can be used for meaningful semantic retrieval of related objects.

\subsection{Embedding with clustering}

We train a convolutional neural network (CNN) to predict foreground and background using oracle labels of patches of  objects and background images. Concurrently, we learn the clustering of objects by imposing constraints that will force the embedding to be partitioned into multiple semantically coherent clusters of objects without explicit labels for different objects.  
 
For our experiments, we use random initialization on the fully-connected layers (the last two layers) and we add the differentiable clustering module after the second to last layer. Note that we only cluster the foreground labels as background activations are not of interest for clustering; the classifier can predict foreground vs background with high accuracy (above 90\%).

The objective function is shown in Equation \ref{eq:kmeans}.
\begin{equation}
L_k = \frac{1}{2N} \sum_{n=1}^N{min_k[(x_n - w_k)^2]}
\label{eq:kmeans}
\end{equation}

In this equation, $N$ is the number of samples, $k$ is the number of defined clusters, $w$ is the “weight” (theoretically and typically the mean of the cluster) for each $k$, and $x$ is the activations from the fully connected layer before the classification fully connected layer.
This is differentiable and the gradient descent algorithm is shown in Equation \ref{eq:derivative}.

\begin{equation}
\delta w_k = w'_k - w_k = \sum_{n=1}^N
\begin{cases}
%\left\{\begin{array}{lr}
l_r(x_n - w_k) & \text{if } k = s(x_n,w)\\
0 & \text{otherwise}
\end{cases}
%\end{array}\right\}
\label{eq:derivative}
\end{equation}

where $s(x_n,w) = argmin_k[x_n]$ and $l_r$ is the learning rate.
We also add $L_2$ regularization over the weights to the loss $L_2=\sum_{j}{w_j^2}$. Furthermore, we use a custom clustering metric $M_C$  that informs us if the clusters are evenly distributed as defined in Equation \ref{eq:reg} and Equation \ref{eq:a}.  This is important since we have no labels to evaluate our data during training.

\begin{equation}
M_C = \frac{1}{NK} \sum_{k=0}^K \sum_{j=k}^K | count_k - count_j |
\label{eq:reg}
\end{equation}

\begin{equation}
count_k = \sum_{n=0}^N
\begin{cases}
0 & \text{if } argmin_k[x_n]=0\\
1 & \text{if } argmin_k[x_n]=1
\end{cases}
\label{eq:a}
\end{equation}

The final loss to be optimized is shown in (Equation~\ref{eq:full_loss})
\begin{equation}
L= L_k + \alpha_r L_2 + \alpha_c L_C
\label{eq:full_loss}
\end{equation}

\noindent where $\alpha_r$ and $\alpha_c$ are hyperparameters which are tuned during the training. $L_C$ is the standard cross-entropy loss for our foreground vs background classification. For our method, we use $\alpha_r = 0.25$ and $\alpha_c = 1$. We apply this loss to every point that is labeled as potentially an object and ignore the background ones when clustering. This way we learn foreground vs background and then learn clustering of the foreground activations. %This is shown in Figure \ref{fig:cluster_diagram}.
Optimization was performed with a `RMSProp' optimizer, with a learning rate of 0.045, momentum 0.9, decay factor 0.9, and $\epsilon$ of 1.0.

%%%%%%%%%%%%%%%%%%%%%%%%%%%%%%%%%%%%%%%%%%%%%%%%%%%%%%%5

\section{Experimental evaluation}

We experiment with a toy example using CIFAR10 and a more challenging example using Cityscapes.

\subsection{CIFAR10 dataset}\label{sec:cifar10}

We first test the proposed unsupervised clustering approach % presented in Section~\ref{sec:approach} 
on the CIFAR10~\cite{krizhevsky2009learning} dataset. The goal of this experiment is to test if clustering can uncover separate categories in a simple toy problem with a two class setting. 

\begin{table} [h]
\centering
\begin{tabular}{ |c|c|c| }
\hline
Clusters & Automobile & Dog\\
\hline
Cluster 0 & 68.5\% & 17.9\%\\
Cluster 1 & 31.5\% & 82.1\%\\

\hline
\end{tabular}
\caption{Unsupervised clustering results on CIFAR10 for discovery of two classes.
Per cluster accuracy for each of the two given classes on the test set (class labels are unknown during training).}
\label{tab:cifar}
\end{table}

We selected as an example the dog and automobile classes to label as foreground. We then train a network from scratch based on the Network in Network architecture (NiN) of \cite{lin2013network} from scratch for our experiments. All other classes of CIFAR are considered background for this experiment. By attaching our modified clustering objective function to the next to last layer, we attempt to cluster dog and automobile without labels.

We can see in our simple experimental results that classes are naturally clustered with the majority of examples correctly assigned. Table \ref{tab:cifar} shows quantitative results on the test set. As seen 68.5\% of the automobile classes and 82.1\% of the dog examples are correctly assigned to separate clusters.
Note that in these cases, the concepts and classes of dog and automobile are unknown to the training algorithm and we are just looking at them after clustering for evaluation.

\subsection{Cityscapes dataset}

The Cityscapes dataset (\cite{cordts2016cityscapes}) is a large-scale dataset that is used for evaluating various classification, detection, and segmentation algorithms related to autonomous driving.
It contains 2975 training, 500 validation, and 1525 test images, where the test set is provided for the purposes of the Cityscape competition only.
In this work, we used the training set for training and the validation set for testing and visualizing results (as the test set has no annotation results). Annotation is provided for classes which represent moving agents in the scene, such as pedestrian, car, motorcycle, bicycle, bus, truck, rider, train. In this work we only use foreground/background labels and intend to discover semantic groups from among the moving objects.

\subsection{Weakly supervised discovery of classes}\label{sec:city}

In this experiment we considered the larger, real-life dataset, Cityscapes (\cite{cordts2016cityscapes}), described above to see if important class categories, e.g. the moving objects in the scene can be clustered into semantically meaningful classes. We extract the locations and extents of the moving objects and use that as weak supervision. Note the classes are uneven and car and person dominate.

We show results clustering 8 categories into 2 and 3 clusters despite the rarity of some of them (such as bicycle).
All results below are presented on the validation set. We report the results in terms of the number of object patches extracted from the available test images.
For this dataset, the CNN architecture is based on the Inception architecture proposed by~\cite{szegedy2015going}. Since there are a small number of examples, we pre-train only the convolutional layers of the network.

Results on clustering the 8 classes of moving objects into 2 and 3 clusters are presented in Table~\ref{tab:kmeans2} and Table~\ref{tab:kmeans3} respectively for the learned embedding by the proposed approach and the baseline embedding. The baseline embedding is calculated by fine-tuning the same architecture in the same manner, but without our loss (Equation~\ref{eq:full_loss}); it uses the same amount of information as input as our embedding. For this experiment, we apply standard kmeans on both activations after training is completed. We see here that our method provides better clustering for the two dominant classes in the dataset (car and person). On the other hand, the baseline embedding clusters on one class only, similar to the two class case. We have consistently observed this behavior for different runs and hypothesize this is due to the sparse nature of the baseline embedding and it's activations.

\begin{figure*}
\includegraphics[width=0.5\textwidth]{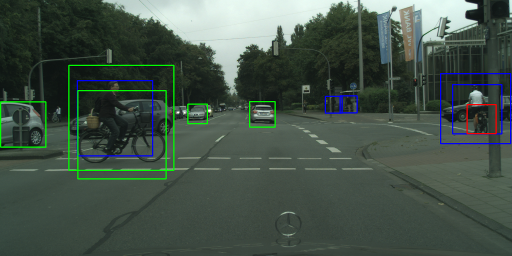}
\includegraphics[width=0.5\textwidth]{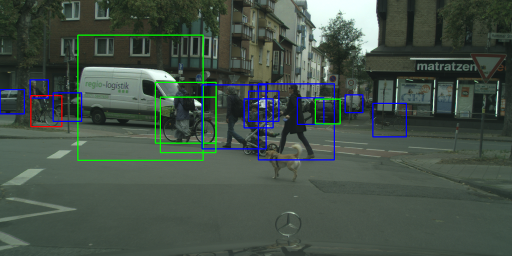}
\includegraphics[width=0.5\textwidth]{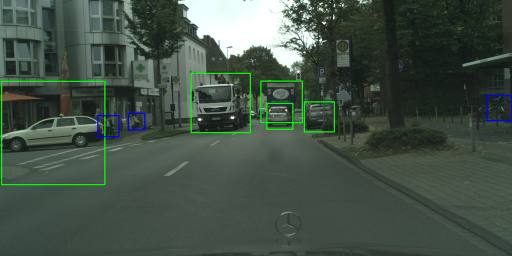}
\includegraphics[width=0.5\textwidth]{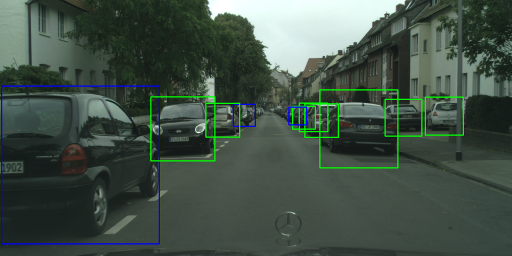}
\includegraphics[width=0.5\textwidth]{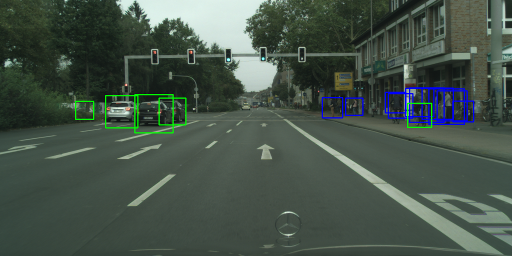}
\includegraphics[width=0.5\textwidth]{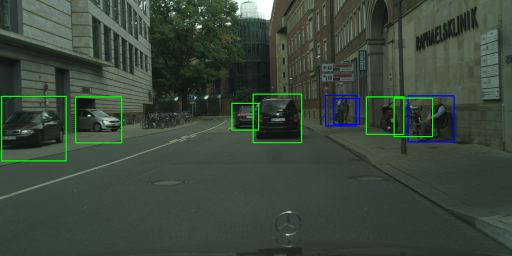}
\includegraphics[width=0.5\textwidth]{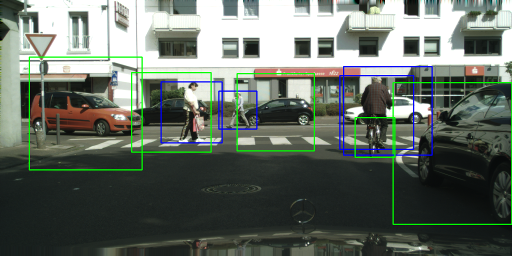}
\includegraphics[width=0.5\textwidth]{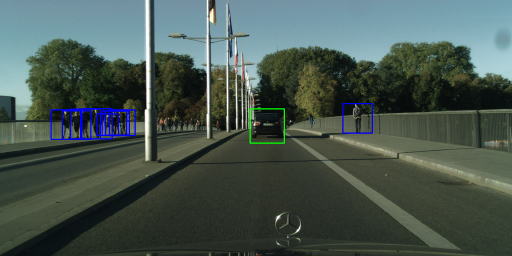}
	\caption{Visualization of clusters learned by our method (for K=3). From the figure, the green class is responsible for retrieving cars, the blue one persons, and the red one bicycles. We can see that both cars and persons are discovered well but bicycles, a rarer class, can be confused with a person or with a partially visible car in the background.}
	\label{fig:cluster_vis}
\end{figure*}

Figure \ref{fig:cluster_vis} visualizes the three retrieved clusters (color-coded) when clustering into 3 clusters with our approach. We can see that people (in blue) and cars (in green) are often correctly retrieved.  Bikes are more rare and may be more often mistaken, for example in cases where a portion of the patch contains part of a car, or since the bicycle very often has a person riding it. Still this is exciting result, given that it is learned by not providing a single class label during training.

\begin{table}
\centering
%\parbox{.5\textwidth}{
\begin{tabular}{ |c|c|c| }
\hline
Classes & Cluster 0 & Cluster 1\\
\hline
Person & 4320 & 138\\
Rider & 676 & 138\\
Car & 1491 & 4399\\
Truck & 60 & 69\\
Bus & 49 & 89\\
Train & 17 & 16\\
Motorcycle & 88 & 205\\
Bicycle & 795 & 787\\
\hline
\end{tabular}

\caption{Unsupervised clustering of objects from Cityscapes using our method. The table shows number of examples assigned to each learned cluster (for K=2).}
\label{tab:kmeans2}
\end{table}

\begin{table}[h]
\centering
\begin{tabular}{ |c|c|c|c||c|c|c| }
\hline
 & \multicolumn{3}{c||}{Our method} & \multicolumn{3}{c|}{Baseline}\\
Classes & Cluster 0 & Cluster 1 & Cluster 2 & Cluster 0 & Cluster 1 & Cluster 2\\
\hline
Person & 151 & 4315 & 17 & 4482 & 1 & 0\\
Rider & 258 & 551 & 7 & 816 & 0 & 0\\
Car & 5195 & 950 & 180 & 6312 & 13 & 0\\
Truck & 89 & 39 & 5 & 131 & 2 & 0\\
Bus & 127 & 20 & 5 & 152 & 0 & 0\\
Train & 25 & 9 & 1 & 35 & 0 & 0\\
Motorcycle & 127 & 76 & 4 & 207 & 0 & 0\\
Bicycle & 1128 & 541 & 450 & 2119 & 0 & 0\\
\hline
\end{tabular}
\caption{Unsupervised clustering on the Cityscapes dataset with 3 clusters. The table shows the number of examples assigned to each learned cluster. Our method (left) and baseline (right). Our method results in 69.98\% accuracy.}
\label{tab:kmeans3}
\end{table}

\section{Conclusions}

We propose a differentiable clustering objective which learns to separate classes during learning and build a better embedding. The key idea is to be able to learn the clusters which are stored as  weights, and simultaneously learn the feature representation and the clustering of the data. Our results show that the proposed approach is useful for extracting semantically related objects. 

\clearpage

{\small
\bibliographystyle{aaai}
\bibliography{egbib}

\begin{thebibliography}{}

\bibitem[\protect\citeauthoryear{Agrawal, Carreira, and
  Malik}{2015}]{agrawal2015learning}
Agrawal, P.; Carreira, J.; and Malik, J.
\newblock 2015.
\newblock Learning to see by moving.
\newblock {\em CVPR}.

\bibitem[\protect\citeauthoryear{Bach and Jordan}{2005}]{bach2003spectral}
Bach, F.~R., and Jordan, M.~I.
\newblock 2005.
\newblock Learning spectral clustering.
\newblock {\em NIPS}.

\bibitem[\protect\citeauthoryear{Barlow}{1989}]{Barlow1989unsupervised}
Barlow, H.
\newblock 1989.
\newblock Unsupervised learning.
\newblock {\em Neural computation}.

\bibitem[\protect\citeauthoryear{Bengio and others}{2009}]{bengio2009learning}
Bengio, Y., et~al.
\newblock 2009.
\newblock Learning deep architectures for ai.
\newblock {\em Foundations and trends{\textregistered} in Machine Learning}
  2(1):1--127.

\bibitem[\protect\citeauthoryear{Bengio, Courville, and
  Vincent}{2012}]{bengio2012unsupervised}
Bengio, Y.; Courville, A.~C.; and Vincent, P.
\newblock 2012.
\newblock Unsupervised feature learning and deep learning: A review and new
  perspectives.
\newblock {\em CoRR, abs/1206.5538}.

\bibitem[\protect\citeauthoryear{Bengio}{2012}]{bengio2012deep}
Bengio, Y.
\newblock 2012.
\newblock Deep learning of representations for unsupervised and transfer
  learning.
\newblock In {\em Proceedings of ICML Workshop on Unsupervised and Transfer
  Learning},  17--36.

\bibitem[\protect\citeauthoryear{Bottou and
  Bengio}{1995}]{bottou1995convergence}
Bottou, L., and Bengio, Y.
\newblock 1995.
\newblock Convergence properties of the k-means algorithms.
\newblock In {\em Advances in neural information processing systems},
  585--592.

\bibitem[\protect\citeauthoryear{Cordts \bgroup et al\mbox.\egroup
  }{2016}]{cordts2016cityscapes}
Cordts, M.; Omran, M.; Ramos, S.; Rehfeld, T.; Enzweiler, M.; Benenson, R.;
  Franke, U.; Roth, S.; and Schiele, B.
\newblock 2016.
\newblock The cityscapes dataset for semantic urban scene understanding.
\newblock In {\em Proceedings of the IEEE Conference on Computer Vision and
  Pattern Recognition},  3213--3223.

\bibitem[\protect\citeauthoryear{Doersch and
  Zisserman}{2017}]{doersch2017multi}
Doersch, C., and Zisserman, A.
\newblock 2017.
\newblock Multi-task self-supervised visual learning.
\newblock {\em arXiv preprint arXiv:1708.07860}.

\bibitem[\protect\citeauthoryear{Erhan \bgroup et al\mbox.\egroup
  }{2010}]{erhan2010does}
Erhan, D.; Bengio, Y.; Courville, A.; Manzagol, P.-A.; Vincent, P.; and Bengio,
  S.
\newblock 2010.
\newblock Why does unsupervised pre-training help deep learning?
\newblock {\em Journal of Machine Learning Research} 11(Feb):625--660.

\bibitem[\protect\citeauthoryear{Faktor and Irani}{2014}]{faktor2014}
Faktor, A., and Irani, M.
\newblock 2014.
\newblock Video segmentation by non-local consensus voting.
\newblock {\em BMVC}.

\bibitem[\protect\citeauthoryear{Grossberg}{1994}]{Grossberg19943d}
Grossberg, S.
\newblock 1994.
\newblock 3-d vision and figure-ground separation by visual cortex.
\newblock {\em Perception and Psychophysics}.

\bibitem[\protect\citeauthoryear{He \bgroup et al\mbox.\egroup
  }{2016}]{he2016deep}
He, K.; Zhang, X.; Ren, S.; and Sun, J.
\newblock 2016.
\newblock Deep residual learning for image recognition.
\newblock {\em CVPR}.

\bibitem[\protect\citeauthoryear{Krizhevsky}{2009}]{krizhevsky2009learning}
Krizhevsky, A.
\newblock 2009.
\newblock Learning multiple layers of features from tiny images.

\bibitem[\protect\citeauthoryear{Kwak \bgroup et al\mbox.\egroup
  }{2015}]{kwak2015unsupervised}
Kwak, S.; Cho, M.; Laptev, I.; Ponce2, J.; and Schmid, C.
\newblock 2015.
\newblock Unsupervised object discovery and tracking in video collections.
\newblock {\em ICCV}.

\bibitem[\protect\citeauthoryear{Lin, Chen, and Yan}{2013}]{lin2013network}
Lin, M.; Chen, Q.; and Yan, S.
\newblock 2013.
\newblock Network in network.
\newblock {\em arXiv preprint arXiv:1312.4400}.

\bibitem[\protect\citeauthoryear{Pathak \bgroup et al\mbox.\egroup
  }{2017}]{pathak2017learning}
Pathak, D.; Girshick, R.; Dollar, P.; Darrell, T.; and Hariharan, B.
\newblock 2017.
\newblock Learning features by watching objects move.
\newblock {\em CVPR}.

\bibitem[\protect\citeauthoryear{Radford, Jozefowicz, and
  Sutskever}{2017}]{radford2017learning}
Radford, A.; Jozefowicz, R.; and Sutskever, I.
\newblock 2017.
\newblock Learning to generate reviews and discovering sentiment.
\newblock {\em arXiv preprint arXiv:1704.01444}.

\bibitem[\protect\citeauthoryear{Ramachandran, Liu, and
  Le}{2016}]{ramachandran2016unsupervised}
Ramachandran, P.; Liu, P.~J.; and Le, Q.~V.
\newblock 2016.
\newblock Unsupervised pretraining for sequence to sequence learning.
\newblock {\em arXiv preprint arXiv:1611.02683}.

\bibitem[\protect\citeauthoryear{Russell \bgroup et al\mbox.\egroup
  }{2006}]{russell2006using}
Russell, B.~C.; Efros, A.~A.; Sivic, J.; Freeman, W.~T.; and Zisserman, A.
\newblock 2006.
\newblock Using multiple segmentations to discover objects and their extent in
  image collections.
\newblock {\em CVPR}.

\bibitem[\protect\citeauthoryear{Singh, Gupta, and
  Efros}{2012}]{singh2012unsupervised}
Singh, S.; Gupta, A.; and Efros, A.~A.
\newblock 2012.
\newblock Unsupervised discovery of mid-level discriminative patches.
\newblock {\em ECCV}.

\bibitem[\protect\citeauthoryear{Sivic \bgroup et al\mbox.\egroup
  }{2005}]{sivic2005discovering}
Sivic, J.; Russell, B.~C.; Efros, A.~A.; Zisserman, A.; and Freeman, W.~T.
\newblock 2005.
\newblock Discovering objects and their location in images.
\newblock {\em ICCV}.

\bibitem[\protect\citeauthoryear{Szegedy \bgroup et al\mbox.\egroup
  }{2015}]{szegedy2015going}
Szegedy, C.; Liu, W.; Jia, Y.; Sermanet, P.; Reed, S.; Anguelov, D.; Erhan, D.;
  Vanhoucke, V.; and Rabinovich, A.
\newblock 2015.
\newblock Going deeper with convolutions.
\newblock {\em CVPR}.

\bibitem[\protect\citeauthoryear{Vijayanarasimhan \bgroup et al\mbox.\egroup
  }{2017}]{sfmnet17}
Vijayanarasimhan, S.; Ricco, S.; Schmid, C.; Sukthankar, R.; and Fragkiadaki,
  K.
\newblock 2017.
\newblock Sfm-net: Learning of structure and motion from video.

\end{thebibliography}
}

\end{document}